\documentclass[letterpaper, 10 pt, conference]{ieeeconf}  

\usepackage{blindtext}
\usepackage{graphicx}
\usepackage{subfigure}
\usepackage{caption}
\usepackage{subfig}
\usepackage{amsmath} 
\usepackage{amssymb}  
\usepackage{bm} 
\usepackage{tikz}
\usepackage{verbatim}
\usepackage{epstopdf}
\usepackage{tabularx}
\usepackage{url}
\usepackage{cite}
\usepackage{booktabs}
\usepackage{multirow}

\IEEEoverridecommandlockouts                              

\title{\LARGE \bf
Laser map aided visual inertial localization in changing environment
}

\author{Xiaqing Ding$^{1}$,
Yue Wang$^{1}$,
Dongxuan Li$^{1}$,
Li Tang$^{1}$,
Huan Yin$^{1}$,
Rong Xiong$^{1}$
\thanks{$^{1}$Xiaqing Ding, Yue Wang, Dongxuan Li, Huan Yin, Li Tang, Rong Xiong are with the State Key Laboratory of Industrial Control and Technology, Zhejiang University, Hangzhou, P.R. China. Yue Wang is the corresponding author {\tt\small wangyue@iipc.zju.edu.cn}. Rong Xiong is the co-corresponding author {\tt\small rxiong@zju.edu.cn}.}%
}

\begin{document}

\maketitle
\thispagestyle{empty}
\pagestyle{empty}

\begin{abstract}

Long-term visual localization in outdoor environment is a challenging problem, especially faced with the cross-seasonal, bi-directional tasks and changing environment. In this paper we propose a novel visual inertial localization framework that localizes against the LiDAR-built map. Based on the geometry information of the laser map, a hybrid bundle adjustment framework is proposed, which estimates the poses of the cameras with respect to the prior laser map as well as optimizes the state variables of the online visual inertial odometry system simultaneously. For more accurate cross-modal data association, the laser map is optimized using  multi-session laser and visual data  to extract the salient and stable subset for localization. To validate the efficiency of the proposed method, we collect data in south part of our campus in different seasons, along the same and opposite-direction route. In all sessions of localization data, our proposed method gives satisfactory results, and shows the superiority of the hybrid bundle adjustment and map optimization.
\end{abstract}


\section{Introduction}

Metric localization in a pre-built map is important for mobile robots to navigate in the environment autonomously. In convention, this problem is mostly addressed by the LiDAR-based map building and monte-carlo localization. However, as the LiDAR is too expensive and heavy to be widely used, the demands for low-cost and light substitutes are risen.

\begin{figure}[]
	\begin{center}
		\includegraphics[width=0.49\textwidth]{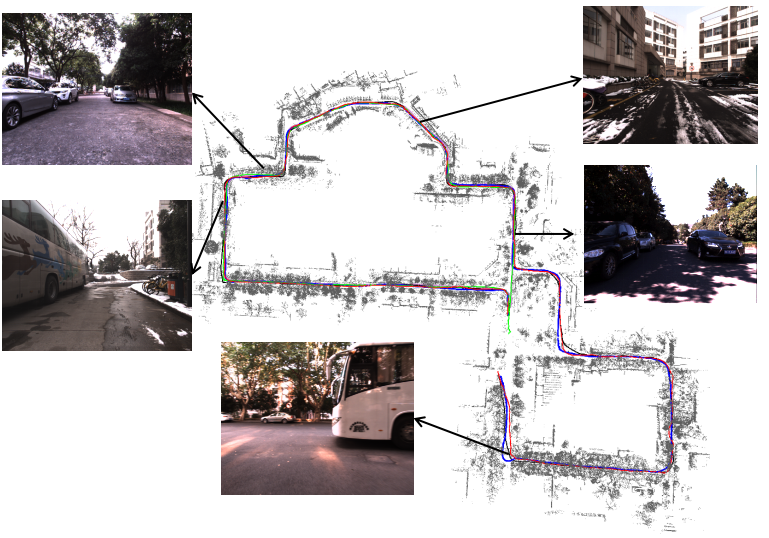}
		\caption{Laser map aided visual inertial localization in changing environment. The map is built using LiDAR in sunny spring, and the lines with different colors are localization trajectories in different sessions across seasons. The five pictures show some challenging circumstances during localization, which include the dynamics, semi-static objects and snow-covered landscape.}
		\label{Overview}
	\end{center}
\end{figure}

In the recent years, camera, because of its low price and versatility, becomes the focus of the robotics community, showing its potential in visual based navigation. Lots of progress has been made on visual inertial navigation system (VINS) or visual inertial simultaneous localization and mapping (VI-SLAM), validating the feasibility of vision-based pose estimation within one session. However, visual localization, which means  localizing the camera in current session on a map that is built in previous sessions, is still challenging since the environment is changing across sessions. In long term, the environmental change involves semi-static objects, like parked cars, and appearance variation, like season and time as shown in Fig. \ref{Overview}. These factors severely deteriorate the precise data association between different sessions, leading to unstable pose estimation. In this paper, we set to study this problem of visual localization across sessions in the changing environment.

Efforts have been paid to address this problem. One way is to combine multiple sessions of map together for localization \cite{churchill2013experience,paton2017expanding}. Thus the features in the current session can find its association in the map with higher probability. The cost of this way is to build a very large map as the variations of the same place should all be saved, thus leading to heavy storage. The other way is to learn the localization based on multi-session data \cite{kendall2015posenet}, which originates from machine learning community. The weakness of this class of methods is that the error is unstable and cannot be explained, hence still calling for further research to guide the practical application.

Recalling the laser-based solution, we can find that the core of its robustness against change in the environment owing to the geometry-based data association and estimation method, which can only be impacted by the semi-static dynamics. Inspired by this clue, the third way of visual localization \cite{Caselitz2016Monocular,Gawel2016Structure,wolcott2014visual,neubertsampling} is promising to solve the problem, which addresses the visual localization by referring to a prior laser map. Generally there are two groups of methods to deal with this problem. One group is to synthesize images from the LiDAR built map \cite{wolcott2014visual,Pascoe2015Robust}. These methods inherently desire lane markings, and the computation is heavy for rendering the image and pose search. The other group is to align the visual map with laser map using some point cloud matching methods like iterative closest point (ICP) \cite{segal2009generalized}. These methods are more general. But the sparse visual map built online can be noisy, and the dense map cannot be built efficiently without intensive computation as pointed in \cite{Gawel2016Structure}. In addition, the uncertainty in the local visual map, especially on depth direction, cannot be considered in ICP, thus introducing more error.

We model the problem of cross-modal localization into the local bundle adjustment (BA) framework to align the sparse visual map to the LiDAR built map. In this framework, the uncertainty in both visual map and the localization between the visual map and LiDAR map are optimized at the same time. Besides, we propose a saliency map extraction method based on multi-session data mining, further improving the localization performance, which also compresses the map storage for faster computation. In overall, the contributions of this paper are addressed below:
\begin{itemize}
	\item Propose a visual inertial localization framework that addresses the localization against the laser map.
	\item Design a hybrid bundle adjustment which optimizes the uncertainty in visual map and the localization at the same time.
	\item Build a compressed laser map by mining the saliency from multi-session laser and visual data.
	\item Evaluate the proposed visual inertial localization method with experiments on multiple sessions of real-world data, which include the cross-seasonal and bi-directional circumstances.
\end{itemize}

The remainder of the paper is organized as follows: In Section II some related works about the visual localization problem are reviewed. Section III gives an overview of the whole visual inertial localization framework that includes an online visual inertial localization system and an offline laser map optimization system, which are introduced in the following Section IV and V. In Section VI we show the experimental results of the proposed visual inertial localization framework evaluated in real-world dataset. And in Section VII some conclusions are addressed.




\section{Related works}


VINS provides precise and high-frequency relative pose estimation along the robot trajectory. Generally, there are two classes of methods in the area, one is to estimate the pose using a nonlinear filter, which is very efficient and light-weighted, thus is preferable in the mobile platform with limited computational resources \cite{mourikis2007multi,wu2015square}. Another class of methods is based on local keyframe-based non-linear optimization, i.e. local bundle adjustment \cite{leutenegger2015keyframe,forster2017manifold}. The optimization based methods can achieve higher performance compared with the filter based solution, but the computational power is relatively higher. The main obstacle preventing VINS from a localization solution is that, VINS has drift in long term, since the accumulated error cannot be eliminated by itself. Therefore, in localization system, VINS is usually employed as a dead reckoning front end to support other back end.

For visual map aided localization back end \cite{schneider2018maplab,lynen2015get,wendel2011natural,ventura2014global}, the maps usually contain the 3D landmarks as well as the corresponding feature-described images. While the robot is tracking with respect to the current local inertial frame, which is the origin of the VINS, the transformation between the local frame and the map frame is computed using some 2D-3D or 2D-2D matching approaches. This matching process can be achieved by either retrieving the similar images from the map dataset \cite{galvez2012bags} or directly retrieving the correspondence with the whole map features \cite{wendel2011natural}. To relieve the computational burden, some approaches for fast retrieving \cite{kendall2015posenet,galvez2012bags} and map compression \cite{dymczyk2015gist,lynen2015get} are studied. These methods enable the localization of the robot in a pre-built visual map, of which the common crucial step is the data association. In long term changing environment, the data association on general feature descriptors cannot give satisfactory performance, thus leading to the failure of the localization.



To relieve the appearance based data association, localizing the robot in a priori LiDAR map with visual sensors is considered as a useful addition \cite{cadena2016past} to compensate for the mutability of visual maps. Since visual and laser information are represented in different modalities, there are generally two ways to align vision-tracked trajectories with the laser map \cite{neubertsampling}. One way \cite{wolcott2014visual,neubertsampling,Pascoe2015Robust} is synthesizing 2D images from the 3D laser map using the intensity or depth information. The currently observed images are matched with the synthesized ones to calculate the relative transformations. To render a high quality image for accurate pose estimation, the LiDAR should be very dense. The accurate pose estimation requires high-resolution search. Thus both steps are highly computational expensive. In addition, this method inherently prefers obvious labels, such as lane markings, thus more appropriate for on-road autonomous driving. The other way \cite{Caselitz2016Monocular,Ozog2016Mapping} is to reconstruct 3D points from the sequential visual images and match them with the priori 3D map to compute the relative transformation between local frame and the laser map frame. Some of these methods apply loosely coupled approaches to align the two kinds of point clouds using point cloud matching methods after visual local bundle adjustment \cite{Caselitz2016Monocular}. But compared with laser point clouds, the reconstructed visual map is sparse and more noisy on depth direction, which cannot be modeled in ICP. In this paper, a hybrid adjustment framework is proposed to achieve the uncertainty reduction in both visual map and the localization. Different from both groups of methods, our method also employ a map compression step to improve the data association and save the storage.

\section{System overview}

Throughout this paper, the local coordinate frame is denoted as $\mathcal{L}$ and the map coordinate frame as $\mathcal{G}$. Meanwhile the current camera frame is denoted as $\mathcal{C}$, laser frame as $\mathcal{F}$ and Inertial Measurement Unit (IMU) frame as $\mathcal{B}$. We represent the pose of the robot as elements of $\mathfrak{se}(3)$, which is denoted as $\mathbf{\xi}\in\mathbb{R}^6$. Also the operators $\mathbf{Exp}(\cdot)$ and $\mathbf{Log}(\cdot)$ as defined in \cite{forster2017manifold} are also included. For each keyframe $i$, the corresponding linear velocity $\mathbf{v_i}\in\mathbb{R}^3$, the IMU bias terms $\mathbf{b_{ai}}\in\mathbb{R}^3$ for acceleration measurement, $\mathbf{b_{gi}}$ for gyroscope measurement together with the pose $\mathbf{\xi_i}$ defined in the local frame $\mathcal{L}$ of the current session, make up the state variable $\mathbf{s_i}=\left\{\mathbf{\xi_i},\mathbf{v_i},\mathbf{b_{ai}},\mathbf{b_{gi}}  \right\}$.

\begin{figure}[]
	\begin{center}
		\includegraphics[width=0.43\textwidth]{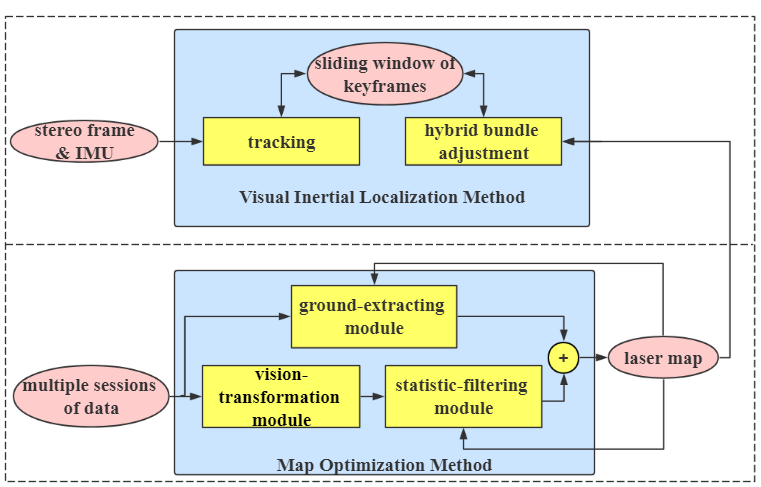}
		\caption{The overview of the laser map aided visual inertial localization framework.}
		\label{mapoptimization}
	\end{center}
\end{figure}

The proposed laser map aided visual inertial localization framework contains an online visual inertial localization method and an offline  map optimization method as shown in Fig. \ref{mapoptimization}. The online localization method tracks the camera pose $\mathbf{\xi_i}$ in the current session using keyframe-based visual inertial odometry using IMU preintegration. To align the trajectory to the map frame, the relative transformation $\mathbf{\xi}^\mathcal{G}_\mathcal{L}$ between the local and the map frame is introduced as "anchor node" \cite{mcdonald20116} which is initialized with a coarse value. When a new keyframe is created, the reconstructed visual points observed by all of the keyframes within the  current sliding-window are transformed into the map frame using the current estimation of $\mathbf{\xi}^\mathcal{G}_\mathcal{L}$ to search their corresponding data association in the laser map respectively. Those matches are designed as constraints which could be added into the local BA along with the variable $\mathbf{\xi}^\mathcal{G}_\mathcal{L}$.  During optimization, the state variables in the sliding-window, visual map points and $\mathbf{\xi}^\mathcal{G}_\mathcal{L}$ are alternatively optimized together to achieve the accurate localization of the robot in the map.


The offline  map optimization method is designed to discover the salient and stable subset within laser maps according to multiple sessions of laser and visual data.
The notations of the maps are defined as follows: The visual map points represented in local frame $\mathcal{L}$ are denoted as $\mathcal{M}^{v,\mathcal{L}}=\left\{ p^v_0,p^v_1...p^v_n\right\}$, where $n$ is the number of map points and the second term in the superscript demonstrates the coordinate frame which the variable is represented in, which could be omitted if there is no ambiguity. The laser map points represented in frame $\mathcal{G}$ are denoted as $\mathcal{M}^l=\left\{ p^l_0,p^l_1...p^l_n\right\}$. The vision-transformed laser map points which are also represented in frame $\mathcal{G}$ are denoted as $\mathcal{M}^{vl}=\left\{ p^{vl}_0,p^{vl}_1... p^{vl}_n\right\}$.
The aim of the map optimization method are fed with $\mathcal{M}^l$ and $\mathcal{M}^{v}$ to build $\mathcal{M}^{vl}$, which is expected to improve the localization accuracy by removing the ambiguous laser points. The map optimization method is a staged filter, consisting of the following three modules as shown in Fig. \ref{mapoptimization}:
\begin{itemize}
\item \textbf{Vision-transformation module}, which removes the irrelevant laser points in each session of map $\mathcal{M}^{l}$ with respect to visual localization problem.
\item \textbf{Statistic-filtering module}, which filters out the dynamic and semi-static points in the vision-transformed map $\mathcal{M}^{vl}$ after accumulating multi sessions of data.
\item \textbf{Ground-extracting module}, which extracts the ground points from each session of data to modify the spatial distribution of the filtered laser map $\mathcal{M}^{vl}$.
\end{itemize}


\section{Visual inertial localization on laser map}

Within the proposed  visual inertial localization method, a keyframe-based visual inertial odometry  with preintegration is applied to track the current pose w.r.t. the local frame $\mathcal{L}$. Meanwhile, a hybrid bundle adjustment  is designed to align the local frame $\mathcal{L}$ to the map frame $\mathcal{G}$ to achieve the localization.

\subsection{Visual inertial odometry}

IMU provides complement measurement in some degenerated situation for vision-based odometry and give observable pitch and roll angles. Generally it works at a much higher frequency than camera. We follow the IMU initialization and keyframe-based visual inertial tracking methods described in \cite{mur2017visual}. The preintegration between two consecutive keyframes are denoted as $[\mathbf{Log}(\Delta \mathbf{R}); \Delta \mathbf{p}; \Delta \mathbf{v}]^T$, where $\mathbf{p}$ and $\mathbf{R}$ are the translation and rotation part of $\mathbf{\xi}$. The discrete evaluation of the preintegration could be derived as \cite{lupton2012visual,forster2017manifold}

\begin{equation}
\begin{split}
\mathbf{R}^{\mathcal{LB}}_{i+1}=&\mathbf{R}^{\mathcal{LB}}_{i}\Delta \mathbf{R}_{i,i+1}\mathbf{Exp}(\mathbf{J}^g_{\Delta \mathbf{R}}\mathbf{b}_{g,i})\\
\mathbf{v}^{\mathcal{B,L}}_{i+1}=&\mathbf{v}^{\mathcal{B,L}}_{i}+\mathbf{g}^{\mathcal{L}}\Delta t_{i,i+1}+\\
&\mathbf{R}^{\mathcal{LB}}_i(\Delta \mathbf{v}_{i,i+1}+\mathbf{J}^g_{\Delta \mathbf{v}}\mathbf{b}_{g,i}+\mathbf{J}^a_{\Delta \mathbf{v}}\mathbf{b}_{a,i})\\
\mathbf{p}^{\mathcal{B,L}}_{i+1}=&\mathbf{p}^{\mathcal{B,L}}_{i}+\mathbf{v}^{\mathcal{B,L}}_{i}\Delta t_{i,i+1}+0.5\mathbf{g}^{\mathcal{L}}\Delta t_{i,i+1}^2+\\
&\mathbf{R}^{\mathcal{LB}}_i(\Delta \mathbf{p}_{i,i+1}+\mathbf{J}^g_{\Delta \mathbf{v}}\mathbf{b}_{g,i}+\mathbf{J}^a_{\Delta \mathbf{v}}\mathbf{b}_{a,i})\\
\end{split}
\end{equation}
where the Jacobian $\mathbf{J}^{g}_{(\cdot)}$ and $\mathbf{J}^a_{(\cdot)}$ represent the first-order approximation of the effect if the variable $(\cdot)$ is changed.

Besides tracking, we keep a sliding-window of the fixed size for bundle adjustment. The cost function could be represented as
\begin{equation}
\mathbf{E}=\sum\limits_{\mathbf{s}_i\in{\mathbf{S}},p^v_j\in{\mathcal{M}^v}}{\mathbf{E}_{ba}(i,j)}+
\sum\limits_{\mathbf{s}_i,\mathbf{s}_k\in{\mathbf{S}}}{\mathbf{E}_{preint}(i,k)}
\label{BA}
\end{equation}
where $\mathbf{S}$ is the set of all of the evaluated states. $\mathbf{E}_{ba}$ represents the reprojection error term between the state $\mathbf{s}_i$ and the visual point $p^v_j$
\begin{equation}
\begin{split}
\mathbf{E}_{ba}(i,j)&=\rho((\pi (p^v_j,\mathbf{\xi}_i)-u_{i,j})^T \mathbf{\Omega_{i,j}}(\pi (p^v_j,\mathbf{\xi}_i)-u_{i,j})) \\
\end{split}
\end{equation}
where $\rho(\cdot)$  is the robust kernel and  $\pi(\cdot,\cdot)$ is the projection function that projects $p^v_j$ onto the image attached to pose $\mathbf{\xi}_i$. $u_{i,j}$ denotes the keypoint matched with the visual point $p^v_j$ on the image attached to pose $\mathbf{\xi}_i$. $\mathbf{\Omega}_{i,j}$ is the information matrix w.r.t. the residual between the projected point and its corresponding keypoint. $\mathbf{E}_{preint}(i,k)$ represents the preintegration error term
\begin{equation}
\begin{split}
	\mathbf{E}_{preint}(i,k)&=\rho([\mathbf{e}^T_R, \mathbf{e}^T_p, \mathbf{e}^T_v]\mathbf{\Omega}_{preint}[\mathbf{e}^T_R, \mathbf{e}^T_p, \mathbf{e}^T_v]^T)\\
	&+\rho(\mathbf{e}^T_b\mathbf{\Omega}_{bias}\mathbf{e}_b)\\
	\mathbf{e}_R=\mathbf{Log}((&\Delta \mathbf{R}_{ik}\mathbf{Exp}(\mathbf{J}^g_{\Delta \mathbf{R}}\mathbf{b}^g_k))^T)\mathbf{R}^{\mathcal{BL}}_i\mathbf{R}^{\mathcal{LB}}_k\\
	\mathbf{e}_p=\mathbf{R}^{\mathcal{BL}}_i(&p^{\mathcal{B,L}}_k-p^{\mathcal{B,L}}_i-\mathbf{v}^{\mathcal{B,L}}_i-0.5\mathbf{g}^{\mathcal{L}}\Delta t^2_{ik})\\
	-(\Delta \mathbf{p}_{ik}&+\mathbf{J}^g_{\Delta p}\mathbf{b}_{gi}+\mathbf{J}^a_{\Delta p}\mathbf{b}_{ak})\\
\mathbf{e}_v=\mathbf{R}^{\mathcal{BL}}_i(&\mathbf{v}^{\mathcal{B,L}}_k-\mathbf{v}^{\mathcal{B,L}}_i-\mathbf{g}^{\mathcal{L}}\Delta t_{ik})-\\
(\Delta \mathbf{v_{ik}}&+\mathbf{J}^g_{\Delta v}\mathbf{b}_{gk}+\mathbf{J}^a_{\Delta v}\mathbf{b}_{ak})\\
\mathbf{e}_b=[\mathbf{b}^T_{ai}, \mathbf{b}&^T_{gi}]^T-[\mathbf{b}^T_{ak}, \mathbf{b}^T_{gk}]^T
\end{split}
\end{equation}
where $\mathbf{\Omega}_{preint}$ is the information matrix of the preintegration and $\mathbf{\Omega}_{bias}$ of the bias.

\subsection{Localization against laser map with hybrid adjustment}

\begin{figure}[]
	\begin{center}
	\includegraphics[width=0.47\textwidth]{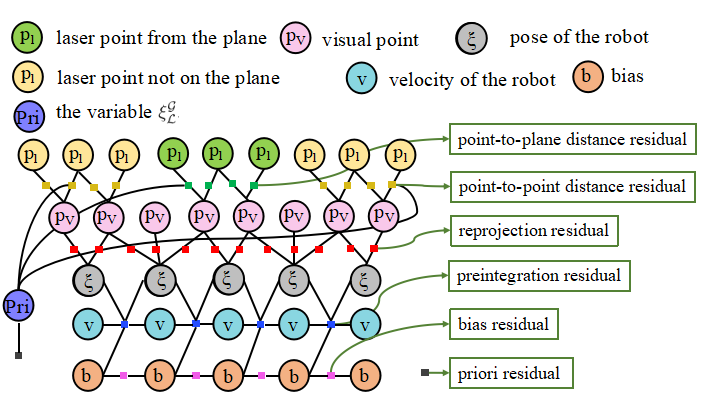}
	\caption{This graph demonstrates the proposed non-rigid bundle adjustment. For representation some lines linked to the variable $\mathbf{\xi}^{\mathcal{G}}_{\mathcal{L}}$ are omitted.}
	\label{graph}
\end{center}
\end{figure}

To align the local frame $\mathcal{L}$ to the map frame $\mathcal{G}$ where the laser map exists, we introduce the relative transformation $\mathbf{\xi}^{\mathcal{G}}_{\mathcal{L}}$ as a variable for estimation and design a hybrid bundle adjustment method which is based on the geometry information of the laser map to eliminate the drift.

At the beginning of the localization, an coarse guess of the relative transformation $\mathbf{\xi}^{\mathcal{G}}_{\mathcal{L}}$ is given for initialization. When a new keyframe is created, all of the visual points $\mathcal{M}^{v}$ observed by the keyframes in the sliding-window are transformed based on the estimation of $\mathbf{\xi}^{\mathcal{G}}_{\mathcal{L}}$ into the map frame. To find correspondence between two point clouds, $k$ nearest neighbors (NN) of each visual point are searched in the laser map $\mathcal{M}^{vl}$. According to the consistency among the normals of the matched laser points , we define two types of the error term corresponding to the point-to-point and point-to-plane metrics. If the normals of the matched laser points are consistent, the error term is designed as the point-to-plane distance error $\mathbf{E}_{pl}$; otherwise the error term will be related to the point-to-point error $\mathbf{E}_{pt}$.
\begin{eqnarray}
\centering
&{r}_{n}(k,j) = (p^{vl}_j-\mathbf{\xi}^{\mathcal{G}}_{\mathcal{L}}(p^v_k))^T \cdot \mathbf{n}_{p^{vl}_j}\\
&\mathbf{E}_{pl}(k,j)=\rho((r_{n}(k,j) n_{p^{vl}_j})^T \Omega_{kj}(r_{n}(k,j) n_{p^{vl}_j})) \\
&\mathbf{E}_{pt}(k,j)=\rho((p^{vl}_j-p^v_k)^T\Omega_{kj}(p^{vl}_j-p^v_k))
\end{eqnarray}
where the $\mathbf{n}_{p^{vl}_j}$ is the normal vector of $p^{vl}_j$.

The relative pose from local to the laser frame estimated in the last step is denoted as $\bar{\mathbf{\xi}^{\mathcal{G}}_{\mathcal{L}}}$, which is introduced as prior to constrain $\mathbf{\xi}^{\mathcal{G}}_{\mathcal{L}}$. The corresponding error term is defined as
\begin{equation}
\mathbf{E}_{prior} =\rho( \mathbf{Log}(\bar{\mathbf{\xi}^{\mathcal{G}}_{\mathcal{L}}}^{-1}\mathbf{\xi}^{\mathcal{G}}_{\mathcal{L}})^T\mathbf{\Omega}_{prior}\mathbf{Log}(\bar{\mathbf{\xi}^{\mathcal{G}}_{\mathcal{L}}}^{-1}\mathbf{\xi}^{\mathcal{G}}_{\mathcal{L}}))
\end{equation}

If we add these error terms into the cost function of bundle adjustment (\ref{BA}) as shown in Fig. \ref{graph}, all of the state variables, visual map points $\mathcal{M}^{v}$ as well as $\mathbf{\xi}^{\mathcal{G}}_{\mathcal{L}}$ could be optimized together, during which the uncertainties of the visual map points and the anchor point  $\mathbf{\xi}^{\mathcal{G}}_{\mathcal{L}}$ are optimized simultaneously aided with the information from the laser map. The cost function is represented as
\begin{equation}
\begin{split}
\mathbf{E}=&\sum{\mathbf{E}_{ba}(i,j)}+
\sum{\mathbf{E}_{preint}(i,k)}\\
&+\sum{\mathbf{E}_{pl}}+\sum{\mathbf{E}_{pt}}+\sum{\mathbf{E}_{prior}}
\end{split}
\end{equation}

In our work, the Cauchy Loss is utilized as the robust kernel and this nonlinear least square optimization problem is solved based on Levenberg-Marquardt algorithm \cite{wright1999numerical}. Since the visual map points $\mathcal{M}^v$ would be adjusted during the optimization, we refer this bundle adjustment as $\textbf{non-rigid bundle adjustment}$.

During non-rigid bundle adjustment, all of the information is utilized for optimization, which leads to high performance of estimation and relieves the accuracy of data association. But if the initial value is not approaching the minima, the whole localization process may result in poor estimation. On the other side, map alignment could also be achieved by applying ICP after the general bundle adjustment, thus the visual map are fixed in this stage. We refer this method as $\textbf{rigid bundle adjustment}$. Note that the method in \cite{Caselitz2016Monocular} belongs to this class, thus is stated in the framework as a special case. Considering the number of variable is reduced as the bias and velocity are separated from localization stage, the rigid bundle adjustment is expected to be more robust to initial value.

To combine the accuracy of the non-rigid bundle adjustment and the robustness to initial value of the rigid bundle adjustment, we also try the third way that the variables are optimized in stage, which is to execute the rigid bundle adjustment first, and then the non-rigid bundle adjustment for higher accuracy.




\section{Prior laser map building and filtering}

For more accurate data association between laser and visual maps, we extract the salient and stable subset from multi-session laser and visual data. To begin with, we model the correspondence between the two modalities.

The localization problem against a given map can be formulated as a maximum likelihood problem:
\begin{equation}\label{loc}
\xi^* = \arg\max \log p(\mathcal{M}^v|\mathcal{M}^l;\xi)
\end{equation}

To demonstrate the data association process, the correspondence variable $c_i$ is introduced which represents that $p^v_i$ is matched with $p^l_j$ if $c_i=j$, thus the likelihood $L$ could be substituted as:
\begin{equation}\label{vpcloc}
L = \sum_i \log \sum_j p(p^v_i,c_i=j|p^l_j;\mathbf{\xi})
\end{equation}
Suppose $Q_i(c_i)$ as a density function of $c_i$, the lower bound of $L$ can be deduced upon Jensen inequality as:
\begin{equation}\label{vpcqlocjensen}
L \geq \sum_i  \sum_j Q_i(c_i) \log \frac{p(p^v_i,c_i=j|p^l_j;\mathbf{\xi})}{Q_i(c_i)}
\end{equation}
This equality only exists if $Q_i(c_i)$ is the posterior of $c_i$
\begin{equation}\label{equa}
Q_i(c_i) = p(c_i=j|p^v_i,p^l_j;\mathbf{\xi})
\end{equation}
Similar to \cite{wang2013based}, we utilize Gaussian distribution to model the likelihood
\begin{equation}\label{likeli}
p(p^v_i|c_i=j,p^l_j;\mathbf{\xi}) = N(p^v_i;R(\mathbf{\xi})p^l_j+t(\mathbf{\xi}),\sigma)
\end{equation}
$R(\mathbf{\xi})$ and $t(\mathbf{\xi})$ represent the rotation and translation components of $\mathbf{\xi}$. If there is no prior on $c_i$, the posterior is
\begin{equation}\label{post}
p(c_i=j|p^v_i,p^l_j;\mathbf{\xi}) = \frac{p(p^v_i|c_i=j,p^l_j;\mathbf{\xi})}{\sum_j p(p^v_i|c_i=j,p^l_j;\mathbf{\xi})}
\end{equation}

The problem described in (\ref{loc}) can be solved with Expectation Maximization (EM) algorithm similar to \cite{bowman2017probabilistic}. If we simplify the posterior as one-peak distribution with the center of the corresponding $j$ which is the maximum of $p(p^v_i|c_i=j,p^l_j;\mathbf{\xi})$, the E-step is equivalent to search the nearest neighbor (NN) of $p^v_i$ in transformed $\mathcal{M}^l$ as correspondence as we utilize in Section IV, which further leads to the solution of EM algorithm of (\ref{loc}) equivalent to ICP algorithm.

To guide the EM to the correct solution, the data association of NN must be close to the correct correspondence. In another word, the posterior (\ref{post}) should be similar with the shape of the finite discrete normal distribution $G(c_i)$ centering on the correct data association
\begin{equation}\label{gt}
G(c_i=j) = N(p^v_i-R(\mathbf{\xi}^*)p^l_j-t(\mathbf{\xi}^*);0,\eta)
\end{equation}
where $\mathbf{\xi}^*$ means the correct pose. We measure the closeness of these two distributions (\ref{post}) and (\ref{gt}) with Kullback-Leibler divergence (KLD)
\begin{equation}\label{kld}
KLD = \sum_j p(c_i=j|p^v_i,p^l_j;\mathbf{\xi})\log\frac{p(c_i=j|p^v_i,p^l_j;\mathbf{\xi})}{G(c_i=j)}
\end{equation}
If we ignore the candidate associations with both very small (\ref{post}) and (\ref{gt}), to reduce (\ref{kld}) we need to increase $p(c_i=j^*|p^v_i,p^l_j;\mathbf{\xi})$ where $j^* = \arg\max G(c_i=j)$, which basically can be achieved following two ways:
\begin{itemize}
	\item \textbf{Pull the correct correspondence}: modify $\mathcal{M}^l$ to make NN of $p^v_i$ in the modified laser map have very large $G(c_i)$, which optimally is equivalent to the $p^l_{j^*}$.
	\item \textbf{Push the wrong correspondence}: modify $\mathcal{M}^l$ to make wrong candidate points in the modified laser map have very small $G(c_i)$.
\end{itemize}

Guided by these two cues, we design a map optimization module which processes the laser map following the above derived theoretical insights to yield $\mathcal{M}^{vl}$.

\subsection{Vision-based map transformation}

Firstly we "pull" the correct correspondence in each data session by  selecting out $p^l_{j^*}$ that is the correct correspondence of $p^v_i$ given the groundtruth pose $\mathbf{\xi}^*$.

To achieve this, we record both the laser and visual information during data collection process. The trajectories are calculated with laser-based SLAM algorithm \cite{pomerleau13comparing} (the first trajectory) or laser-based localization algorithm (the following trajectories), thus all sessions of collected data are represented in the same map coordinate frame $\mathcal{G}$. Transform each laser scan into its corresponding camera frame then project the laser points onto the image
\begin{equation}\label{reprojection}
u^l_j = \pi(R(\mathbf{\xi}^{\mathcal{C}}_{\mathcal{F}})p^{l}_j + t(\mathbf{\xi}^{\mathcal{C}}_{\mathcal{F}}))
\end{equation}
where $\pi(\cdot)$ represents the reprojection function and $\mathbf{\xi}^{\mathcal{C}}_{\mathcal{F}}$ means the extrinsic transformation between camera and laser coordinate frames.

We extract feature points on each image using the same type of extractor which will be used in visual inertial system. Then among the extracted feature points  we search the closet point $u^v_j$ for each $u^l_j$. If the distance between them satisfies
\begin{equation}
D(u^v_i,u^l_j) < \delta_u
\end{equation}
where $D(\cdot,\cdot)$ returns the distance between the input points, then $u^l_j$ is considered as a correct correspondence of $u^v_j$ and is inserted into the vision-transformed map of the corresponding session. When all of the collected sessions have been processed, these vision-transformed maps $\left\{\mathcal{M}^{vl}_i\right\}_{i=0}^{n-1}$ are delivered into statistic-filtering module for further optimization.

\subsection{Static map maintenance}

To further ``push'' away the wrong correspondence, we select only the stable laser points out of multi sessions of data to avoid wrong data association which might be introduced by the dynamics.

Initially, we keep all of the points in $\mathcal{M}^{vl}_0$ as the base map $\mathcal{M}^{vl}$. For each laser point $p^{vl,\mathcal{M}^{vl}_i}_j$ in map $\mathcal{M}^{vl}_i$, we search its nearest neighbor $p^{vl,\mathcal{M}^{vl}}_k$ in base map $\mathcal{M}^{vl}$. If the distance satisfies
\begin{equation}\label{distance map maintenance}
D(p^{vl,M^{vl}_i}_j,p^{vl,M^{vl}}_k) > d^\alpha
\end{equation}
$p^{vl,M^{vl}_i}_j$ should be collected from the new part of the environment that has never been observed in the past processed $i$ sessions of data. Otherwise the matched point $p^{vl,M^{vl}}_k$ is considered as being observed another time.

After all of the collected sessions have been processed, the points in the merged base map $\mathcal{M}^{vl}$ should cover most part of the environment, dynamics included. We denote the number that each laser point has been observed as "observation number", which could be utilized as an index to determine whether a laser point belongs to the static part of the environment or not
\begin{equation}\label{part}
\mathcal{M}^{vl} = \left\{\begin{array}{cc}
\mathcal{M}^{vl}_{s}, & obs(p^{vl,\mathcal{M}^{vl}_{s}}_i)\geq\beta \\
\mathcal{M}^{vl}_{d}, & obs(p^{vl,\mathcal{M}^{vl}_{d}}_i)<\beta
\end{array}
\right.
\end{equation}
where $obs(\cdot)$ returns the observation number of the input point. $\mathcal{M}^{vl}_s$ represents the subset in $\mathcal{M}^{vl}$ in which the points belong to the static part of the environment, and $\mathcal{M}^{vl}_d$ represents the dynamic part.

Since we do not apply any model assumption when classify the map points, we further apply two post-processing steps to discover the false positive and false negative classification results which are introduced by overlapping. To filter out the false positive points in map $\mathcal{M}^{vl}_{s}$, we search the nearest point $p^{vl,\mathcal{M}^{vl}_{d}}_j$ in map $\mathcal{M}^{vl}_{d}$ for  each point $p^{vl,\mathcal{M}^{vl}_{s}}_i$ then apply an erosion method to filter out those points in  $\mathcal{M}^{vl}_{s}$ that satisfy
\begin{equation}
\left\{ \begin{array}{c}
D(p^{vl,{M}^{vl}_{s}}_i,p^{vl,{M}^{vl}_{d}}_j)< d^{er}\\
obs(p^{vl,\mathcal{M}^{vl}_{s}}_i)<\beta^{er}
\end{array}
\right.
\end{equation}

Next, to fetch the false negative points which have been classified into map $\mathcal{M}^{vl}_{d}$,  we search the nearest point $p^{vl,\mathcal{M}^{vl}_{s}}_j$ for each point $p^{vl,\mathcal{M}^{vl}_{d}}_i$ in map $\mathcal{M}^{vl}_{s}$ then apply an expansion method to select those points in  $\mathcal{M}^{vl}_{d}$ which satisfy
\begin{equation}
\left\{ \begin{array}{c}
D(p^{vl,{M}^{vl}_{d}}_i,p^{vl,{M}^{vl}_{s}}_j)< d^{ep}\\
obs(p^{vl,\mathcal{M}^{vl}_{d}}_i)>\beta^{ep}
\end{array}
\right.
\end{equation}
then add them back into map $\mathcal{M}^{vl}_{s}$.
After the post-processing steps, most of the map $\mathcal{M}^{vl}_{s}$ should belong to the static and vision-related subset of the original laser map.

\subsection{Ground points modification}

When extracting feature points on an image, some feature points might be extracted from the textureless area or edges of shadow on the ground. These points are highly sensitive to the illumination, which means that they are hardly extracted at the same position at another time, thus are excluded from the map $\mathcal{M}^{vl}_{s}$. However, the ground feature points in the current session can still be extracted. As a result, the NN based search may encourage these points to match the map, which must leads to the wrong data association, especially when the initial pose is noisy.

To improve the robustness of data association when the initial pose is noisy, we assign the ground points with the point-to-plane error term, as the ground is locally planar. The point-to-plane error term only care about the perpendicular distance, thus any data association between points on the local ground can be regarded correct. We select the laser points from the ground in each scan of data based on  height filter as the position of the LiDAR with respect to the robot is considered as known and fixed. All of the ground points are merged and further filtered by a voxel filter into the ground map. Finally, the filtered map $\mathcal{M}^{vl}$ and the sampled ground map  are merged into the final cross-modal stable map $\mathcal{M}^{vl}$ for visual inertial localization.

\section{Experimental results}

\begin{figure}[]
	\begin{center}
		\includegraphics[width=0.47\textwidth]{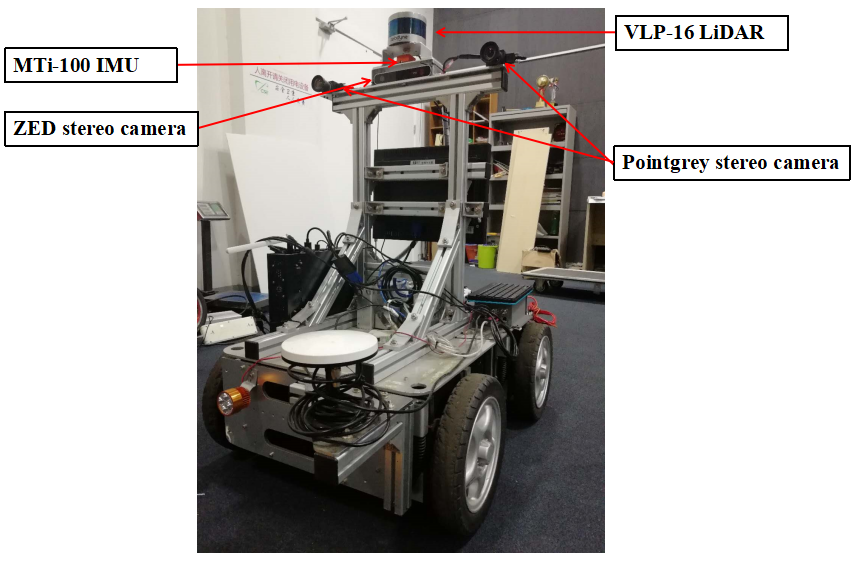}
		\caption{The equipment used for collecting data.}
		\label{equipment}
	\end{center}
\end{figure}

\begin{table}
	\centering
	\caption{Overview of YQ South datasets}
	\label{tab:dataset}
	\begin{tabular}{cccc}
		\hline
		 \multicolumn{4}{c}{Map Building Dataset}\\
		 Start Time & Duration & Start Time & Duration  \\
		
		 2017/03/03 07:52:31 & 17:44 & 2017/03/03 09:20:13 & 18:45  \\
		 2017/03/03 10:23:11 & 18:14 & 2017/03/03 11:48:03 & 18:17  \\
		 2017/03/03 12:59:16 & 19:12 & 2017/03/03 14:34:43 & 19:24  \\
		 2017/03/03 16:05:54 & 18:39 & 2017/03/03 17:38:14 & 18:01  \\
		 2017/03/07 07:43:30 & 17:54 & 2017/03/07 09:06:04 & 18:46  \\
		 2017/03/07 10:19:45 & 19:04 & 2017/03/07 12:40:29 & 18:42  \\
		 2017/03/07 14:35:16 & 19:01 & 2017/03/07 16:28:26 & 17:59 \\
		 2017/03/07 17:25:06 & 18:34 & 2017/03/07 18:07:21 & 19:49 \\
		 2017/03/09 09:06:05 & 17:50 & 2017/03/09 10:03:57 & 17:52 \\
		 2017/03/09 11:25:40 & 18:17 & 2017/03/09 15:06:14 & 19:13 \\
		 2017/03/09 16:31:34 & 19:36 &  & \\
		\hline
		\multicolumn{4}{c}{Testing Dataset}\\
		Start Time & Duration & Start Time & Duration  \\
		2017/08/23 09:40:13 & 16:31 & 2017/08/24 09:21:41 & 13:21 \\
		2017/08/27 15:22:11 & 17:03 & 2017/08/28 17:06:06 & 17:15 \\
		2018/01/29 11:09:15 & 14:59 \\
		\hline
		
	\end{tabular}
\end{table}

To evaluate the performance of the proposed vision-related laser map optimization method and the laser map aided visual inertial localization method, we collect data in southern Yuquan Campus, Zhejiang University, China. The experimental platform to collect data for map construction is a four-wheel mobile robot equipped with a VLP-16 Velodyne LiDAR, a MTi 100 IMU and a ZED stereo camera. 
When collect data for evaluating the localization method, the vision sensor is replaced by a Pointgrey stereo camera. The equipment is shown in Fig. \ref{equipment}.
All the methods are deployed on a computer with an Intel i7-6700 CPU 3.40GHz and 16G RAM using C++. The ICP algorithm used in our method is implemented based on libpointmatcher \cite{pomerleau13comparing}. We extract ORB features during localization referring to \cite{mur2015orb} and solve the nonlinear optimization problem with Ceres \cite{ceres-solver}.

\subsection{YQ datasets overview}


The dataset used for map building includes 21 sessions of data collected from three days in March, 2017, which are listed detailedly in Table \ref{tab:dataset}. The robot is under remote control and each session of data is collected almost along the same way with the length of around 1,300 meters as shown in Fig. \ref{Overview}. As a campus environment, the high-dynamics include pedestrians, cyclists, moving cars and buses. While the low-dynamics include parking cars, the shape of the trees, the landmarks on the road and products of the weather like snow.

The dataset for testing the visual inertial localization method contains four sessions of data collected in summer and one session of data collected in winter as shown in Table \ref{tab:dataset}, in which a session of summer data (2017/08/24) is collected along the opposite way of the others to test the bi-directional performance. Besides, the other three sessions of data are collected at the different time of a day. And the last session of data is collected after snow where the snow extremely changes the visual features on the road.

We utilize the laser-based localization method \cite{pomerleau13comparing} to evaluate the trajectories for each session of data as the groundtruth. The map used for laser-based localization is constructed using the firstly collected session of data. Thus all of the groundtruth is represented in the same coordination frame.

\subsection{Map construction}

\begin{figure}[]
	\begin{center}
		\includegraphics[width=0.5\textwidth]{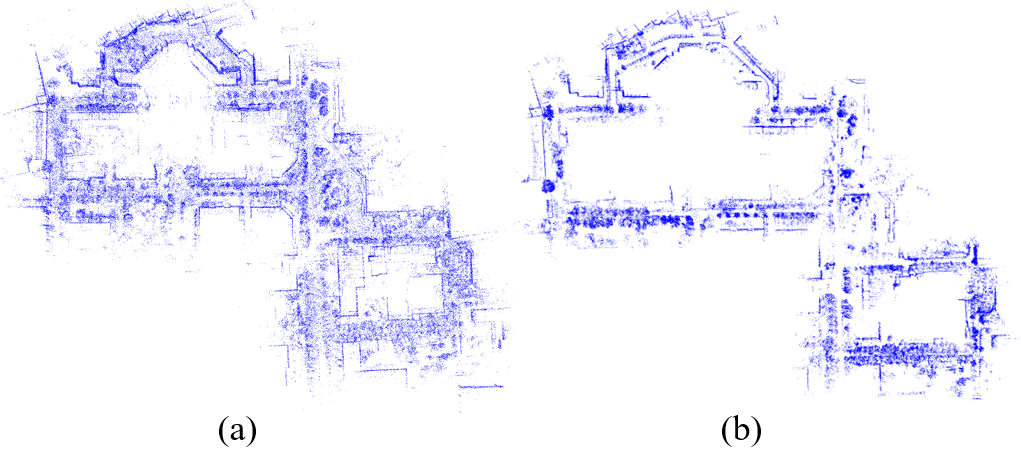}
		\caption{The full laser map (a) and the filtered laser map (b).}
		\label{comparemap}
	\end{center}
\end{figure}

We utilize 21 sessions of data collected during three days in March, 2017.
All of the sessions are merged using the method in Section V. The final extracted map is shown in Fig. \ref{comparemap} (b). The map shown in Fig. \ref{comparemap} (a) is constructed with laser-based SLAM method \cite{pomerleau2014long} using the first session of data. It's easy to figure out that the spatial distributions of these two maps are different and the spatial distribution of the filtered map is more like the one of traditional visual constructed map.

\subsection{Evaluations of localization}

\begin{figure}[htb]
\centering
		\includegraphics[width=8cm]{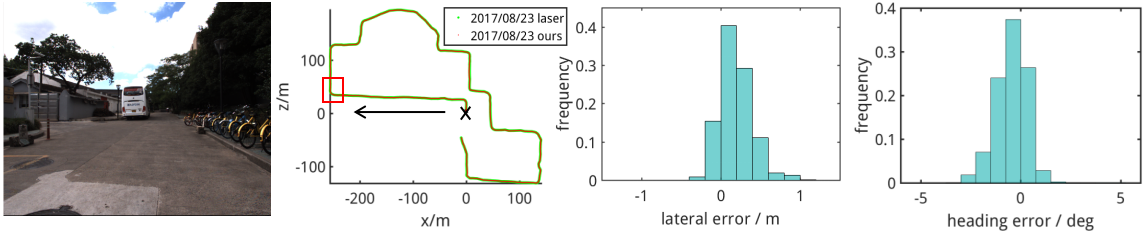}
		\includegraphics[width=8cm]{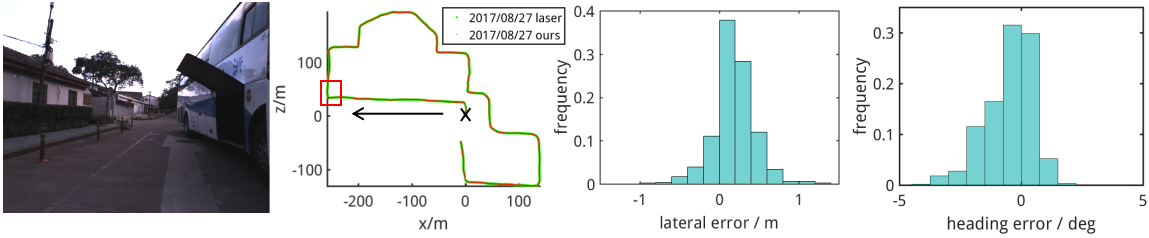}
		\includegraphics[width=8cm]{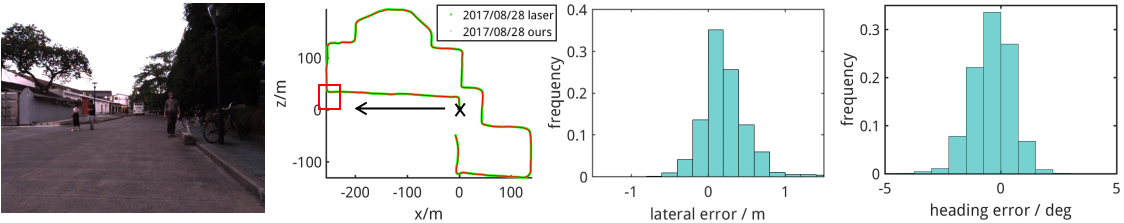}
		\includegraphics[width=8cm]{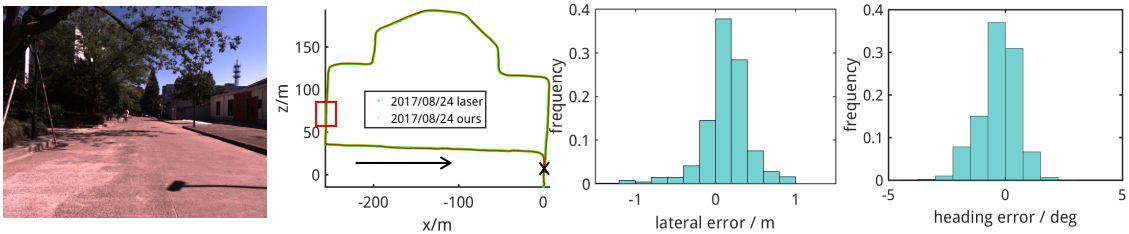}
		\includegraphics[width=8cm]{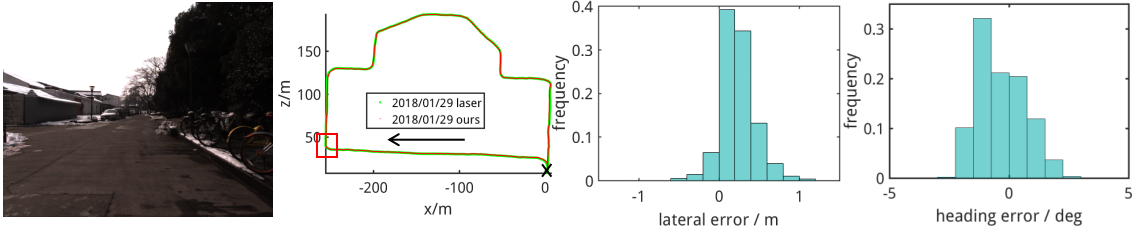}
	\caption{Localization trajectories (second column) and histograms of lateral (third column) and heading (fourth column) errors in YQ South datasets. The black crosses indicate the beginning of the trajectories and the arrows indicate the directions of the trajectories. The pictures in the first column show the views observed  in different sessions at the positions that labeled in the second column with red rectangles.}
	\label{yq_localization}
\end{figure}

To validate the efficiency of the proposed localization method, we evaluate it in the five sessions of data collected in summer and winter. We apply the hybrid bundle adjustment at different frequency of the non-rigid bundle adjustment. We demonstrate the frequency as the proportion of executed times. For example, "1:3" means after one time of non-rigid bundle adjustment three times of rigid bundle adjustment will be executed. Also the performance of totally applying the non-rigid bundle adjustment and rigid bundle adjustment are also evaluated. The results are shown in Table \ref{tab:ate}, inside the bold numbers  represent the best results for each session of data, and the corresponding trajectories as well as the lateral and heading errors are shown in Fig. \ref{yq_localization}.

\begin{table}[htbp]
	\centering
	\caption{The results of the localization method evaluated with the mean value of the Absolute Trajectory Error (ATE)}
	\begin{tabular}{cccccc}
		\toprule
		Sequences  & non-rigid & 1:1   & 1:3   & 1:5   & rigid \\
		\midrule
		23/08/2017 09:40:13 & 0.580  & 0.473  & 0.347  & \textbf{0.315} & 0.792 \\
		24/08/2017 09:21:41 & 0.662  & 0.494  & 0.484  & \textbf{0.447} & 1.015 \\
		27/08/2017 15:22:11 & 0.956  & \textbf{0.417} & 0.468  & 0.480  & 0.944 \\
		28/08/2017 17:06:06 & 0.582  & 0.434  & \textbf{0.392} & 0.440  & 0.721 \\
		29/01/2018 11:09:15 & 0.683  & 0.429  & 0.435  & \textbf{0.391} & 0.605 \\
		average & 0.693 & 0.449 & 0.425 & \textbf{0.415} & 0.815\\
		\bottomrule
	\end{tabular}%
	\label{tab:ate}%
\end{table}%

As the results show, the hybrid method outperforms the non-rigid and rigid methods in every session of data. Averagely, the proportion "1:5" gives the best performance.

\subsection{Evaluations of map optimization }
For comparison, we also evaluate the performance of  localizing  within the full map that is not filtered by the proposed map optimization method. The results are listed in Table \ref{tab:compare}. The hybrid method for full map aided localization utilizes the same frequency at which the filtered map aided method gives the best results.

From the table, we can see that in some sequences, the results of full map aided method are almost the same with the filtered map aided method. But in other sequences their results differ a lot, especially in the session collected in 23, Aug, 2017 where the rigid bundle adjustment based method fails to run over the whole trajectory within the full map. This validates the efficiency of the map optimization method for long-term localization tasks.

\begin{table}[htbp]
	\centering
	\caption{comparison of the localization performance in different maps}
	\begin{tabular}{ccccc}
		\toprule
		\multicolumn{1}{c}{\multirow{2}[2]{*}{Sequences}} & \multicolumn{2}{c}{filtered map} & \multicolumn{2}{c}{full map} \\
		\multicolumn{1}{c}{} & \multicolumn{1}{l}{hybrid} & \multicolumn{1}{l}{rigid} & hybrid & rigid \\
		\midrule
		23/08/2017 09:40:13 & \textbf{0.315} & 0.792  & {0.847} & / \\
		24/08/2017 09:21:41 & \textbf{0.447} & 1.015  & 0.459  & {0.451} \\
		27/08/2017 15:22:11 & \textbf{0.417} & 0.944  & {0.509} & 0.564 \\
		28/08/2017 17:06:06 & \textbf{0.392} & 0.721  & 1.879  & {0.539} \\
		29/01/2018 11:09:15 &\textbf{0.391} & 0.605 & 0.536  & {0.408} \\
		average & \textbf{0.415} & 0.815 & 0.843 & {0.488} \\
		\bottomrule
	\end{tabular}%
	\label{tab:compare}%
\end{table}%

\section{Conclusions}

In this paper, the cross-modal localization problem is addressed based on a hybrid bundle adjustment framework to align the sparse visual map to the LiDAR built map. Both the uncertainties of the visual map and the localization estimation are optimized simultaneously. To improve the accuracy of data association between the sparse visual map and dense laser map, a saliency map extraction method is proposed which also filters out the dynamics in the laser data. The efficiency of this proposed method is validated with experiments on multiple sessions of real-world data, which include the cross-seasonal and bi-directional circumstances.

In the future, we would like to learn the map extraction method proposed in this paper to filter the laser maps of the new environment. In addition, the visual SLAM aided by the prior laser map is also our potential research direction.

\section*{Acknowledgement}

This work was supported in part by the National Nature Science Foundation of China under Grant U1609210 and Grant 61473258, and in part by the Science Fund for Creative Research Groups of NSFC under Grant 61621002.

\addtolength{\textheight}{-12cm}   




\bibliographystyle{IEEEtran}
\bibliography{library}

\end{document}